\documentclass[11pt,a4paper]{article}
\usepackage[hyperref]{acl2019}
\usepackage{times}
\usepackage{latexsym}
\usepackage{url}
\usepackage{amsmath}
\usepackage{amssymb}
\usepackage{hyperref}

\usepackage{graphicx}

\newcommand{\compactpara}[1]{\noindent\textbf{#1}}

\aclfinalcopy 
\def\aclpaperid{903} 

\makeatletter
\newcommand{\printfnsymbol}[1]{%
  \textsuperscript{\@fnsymbol{#1}}%
}
\makeatother

\title{A Brief Survey of Multilingual Neural Machine Translation}

\author{Raj Dabre\thanks{equal contribution} \\
    NICT \\
  Kyoto, Japan \\
  {\tt raj.dabre@nict.go.jp} \\ \And
  Chenhui Chu\printfnsymbol{1} \\
  Institute for Datability Science \\
  Osaka University \\
  Osaka, Japan. \\
  {\tt chu@ids.osaka-u.ac.jp} \\ \And
  Anoop Kunchukuttan\printfnsymbol{1} \\
  Microsoft AI \& Research, \\
  Hyderabad, India. \\
  {\tt ankunchu@microsoft.com} \\  
  }

\date{}

\begin{document}
\maketitle
\begin{abstract}
We present a survey on multilingual neural machine translation (MNMT), which has gained a lot of traction in the recent years. MNMT has been useful in improving translation quality as a result of knowledge transfer. MNMT is more promising and interesting than its statistical machine translation counterpart because end-to-end modeling and distributed representations open new avenues. Many approaches have been proposed in order to exploit multilingual parallel corpora for improving translation quality. However, the lack of a comprehensive survey makes it difficult to determine which approaches are promising and hence deserve further exploration. In this paper, we present an in-depth survey of existing literature on MNMT. We categorize various approaches based on the resource scenarios as well as underlying modeling principles. We hope this paper will serve as a starting point for researchers and engineers interested in MNMT.
\end{abstract}

\section{Introduction}

Neural machine translation (NMT) 
\cite{DBLP:journals/corr/ChoMBB14,DBLP:journals/corr/SutskeverVL14,bahdanau15} has become the dominant paradigm for MT in academic research as well as commercial use \cite{DBLP:journals/corr/WuSCLNMKCGMKSJL16}. NMT has shown state-of-the-art performance for many language pairs \cite{bojar-EtAl:2017:WMT1,bojar2018wmtfindings}. Its success can be mainly attributed to the use of distributed representations of language, enabling end-to-end training of an MT system. Unlike statistical machine translation (SMT) systems \cite{koehn-EtAl:2007:PosterDemo}, separate lossy components like word aligners, translation rule extractors and other feature extractors are not required. 
The dominant NMT approach is the \textit{Embed - Encode - Attend - Decode} paradigm. Recurrent neural network (RNN) \cite{bahdanau15}, convolutional neural network (CNN) \cite{gehring17} and self-attention \cite{NIPS2017_7181} architectures are popular approaches based on this paradigm. For a more detailed exposition of NMT, we refer readers to some prominent tutorials \cite{neubig17nmt,koehn17nmt}.


While initial research on NMT started with building translation systems between two languages, researchers discovered that the NMT framework can naturally incorporate multiple languages. Hence, there has been a massive increase in work on MT systems that involve more than two languages \cite{dong15,firat16,N16-1004,ijcai2017-555,johnson17,P17-1176,DBLP:conf/aaai/ChenLL18,D18-1103} \textit{etc}. We refer to NMT systems handling translation between more than one language pair as \textit{multilingual NMT} (MNMT) systems. The ultimate goal MNMT research is to develop one model for translation between all possible languages by effective use of available linguistic resources.

\begin{figure*}
    \centering
    \includegraphics[width=\textwidth]{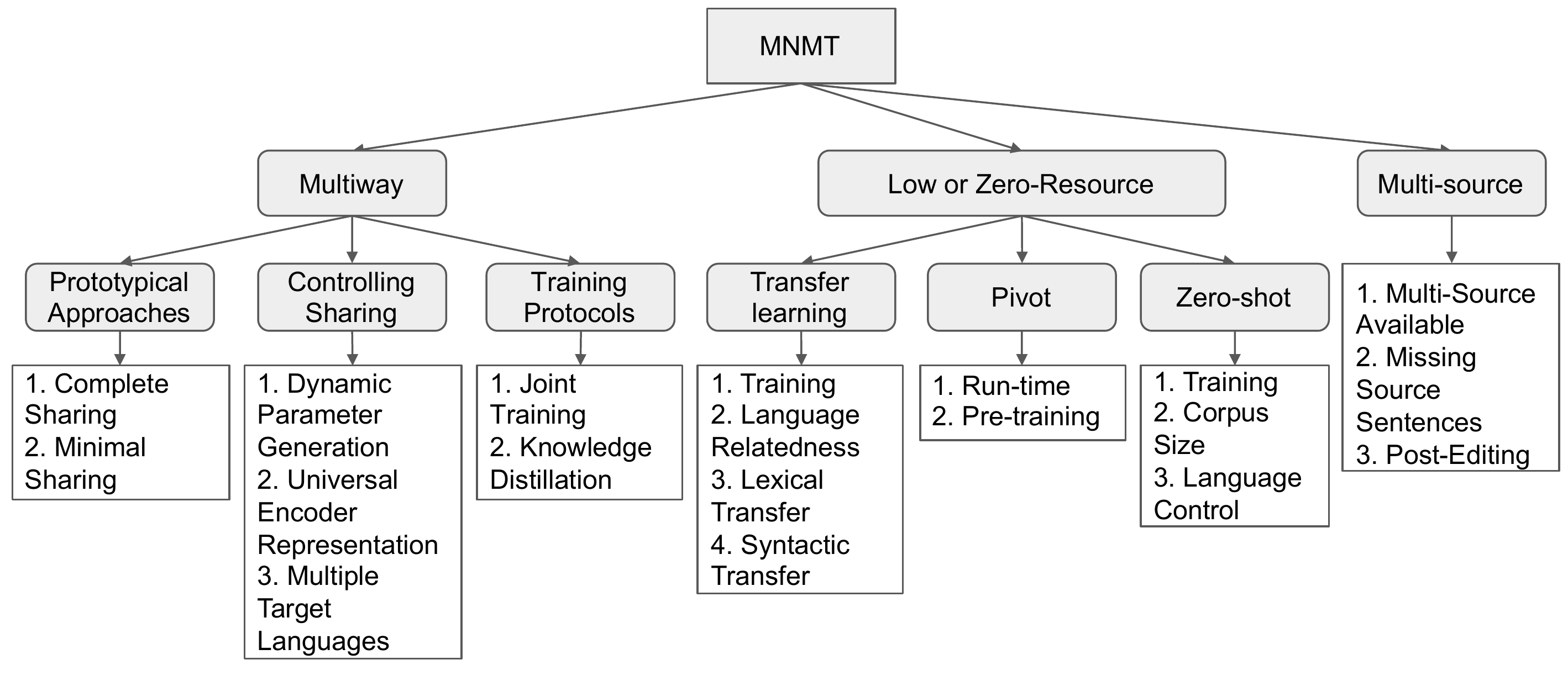}
    \vspace{-6mm}
    \caption{MNMT research categorized according to resource scenarios and underlying modeling principles.}
    \vspace{-4mm}
        \label{fig:overview}
\end{figure*}

MNMT systems are desirable because training models with data from many language pairs might help acquire knowledge from multiple sources \cite{N16-1004}. Moreover, MNMT systems tend to generalize better due to exposure to diverse languages, leading to improved translation quality. This particular phenomenon is known as knowledge transfer \cite{Pan:2010:STL:1850483.1850545}. 
Knowledge transfer has been strongly observed for translation between low-resource languages, which have scarce parallel corpora or other linguistic resources but have benefited from data in other languages \cite{DBLP:conf/emnlp/ZophYMK16:original}. In addition, MNMT systems will be compact, because a single model handles translations for multiple languages \cite{johnson17}. This can reduce the deployment footprint, which is crucial for constrained environment like mobile phones or IoT devices. It can also simplify the large-scale deployment of MT systems. 
Most importantly, we believe that the biggest benefit of doing MNMT research is getting better insights into and answers to an important question in natural language processing: \textit{how do we build distributed representations such that similar text across languages have similar representations?}

There are multiple MNMT scenarios based on available resources and studies have been conducted for the following scenarios (Figure \ref{fig:overview}\footnote{Please see the supplementary material for papers related to each category.}): 

\compactpara{Multiway Translation.} The goal is constructing a single NMT system for one-to-many \cite{dong15}, many-to-one \cite{lee17} or many-to-many \cite{firat16} translation using parallel corpora for more than one language pair.

\compactpara{Low or Zero-Resource Translation.} For most of the language pairs in the world, there are small or no parallel corpora, and three main directions have been studied for this scenario.
{\it Transfer learning}: Transferring translation knowledge from a high-resource language pair to improve the translation of a low-resource language pair \cite{DBLP:conf/emnlp/ZophYMK16:original}.
{\it Pivot translation}: Using a high-resource language (usually English) as a pivot to translate between a language pair \cite{firat16}.
{\it Zero-shot translation}: Translating between language pairs without parallel corpora \cite{johnson17}. 

\compactpara{Multi-Source Translation.} Documents that have been translated into more than one language might, in the future, be required to be translated into another language. In this scenario, existing multilingual redundancy in the source side can be exploited for multi-source translation \cite{N16-1004}.

Given these benefits, scenarios and the tremendous increase in the work on MNMT in recent years, we undertake this survey paper on MNMT to systematically organize the work in this area. To the best of our knowledge, no such comprehensive survey on MNMT exists. Our goal is to shed light on various MNMT scenarios, fundamental questions in MNMT, basic principles, architectures, and datasets of MNMT systems.
The remainder of this paper is structured as follows:  
We present a systematic categorization of different approaches to MNMT in each of the above mentioned scenarios to help understand the array of design choices available while building MNMT systems (Sections \ref{sec:multiway}, \ref{sec:low}, and \ref{sec:multisource}). We put the work in MNMT into a historical perspective with respect to multilingual MT in older MT paradigms (Section \ref{sec:history}). We also describe popular multilingual datasets and the shared tasks that focus on multilingualism (Section \ref{sec:data}). In addition, we compare MNMT with domain adaptation for NMT, which tackles the problem of improving low-resource in-domain translation (Section \ref{sec:domain}).
Finally, we share our opinions on future research directions in MNMT (Section \ref{sec:future}) and conclude this paper (Section \ref{sec:conlusion}).

\section{Multiway NMT}
\label{sec:multiway}

The goal is learning a single model for $l$ language pairs $(s_i,t_i) \in \mathbf{L}$ ~ $(i=1 \textrm{ to } l)$, where $\mathbf{L} \subset S \times T$, and $S, T$ are sets of source and target languages respectively. $S$ and $T$ need not be mutually exclusive. Parallel corpora are available for these $l$ language pairs. One-many,  many-one and many-many NMT models have been explored in this framework.
Multiway translation systems follow the standard paradigm in popular NMT systems. However, this architecture is adapted to support multiple languages. The wide ranges of possible architectural choices is exemplified by two highly contrasting prototypical approaches. 

\vspace{-2mm}
\subsection{Prototypical Approaches}
\vspace{-1mm}

\compactpara{Complete Sharing}. \citet{johnson17} proposed a highly compact model where all languages share the same embeddings, encoder, decoder, and attention mechanism. A common vocabulary, typically subword-level like byte pair encoding (BPE) \cite{DBLP:journals/corr/SennrichHB15}, is defined across all languages. The input sequence includes a special token (called the \textit{language tag}) to indicate the target language. This enables the decoder to correctly generate the target language, though all target languages share the same decoder parameters. The model has minimal parameter size as all languages share the same parameters; and achieves comparable/better results w.r.t. bilingual systems. But, a massively multilingual system can run into capacity bottlenecks \cite{aharoni19}. This is a \textit{black-box} model, which can use an off-the-shelf NMT system to train a multilingual system. \citet{ha16} proposed a similar model, but they maintained different vocabularies for each language. 

This architecture is particularly useful for related languages, because they have high degree of lexical and syntactic similarity \cite{sachan18}. Lexical similarity can be further utilized by (a) representing all languages in a common script using script conversion \cite{dabre18wat,lee17} or transliteration (\citet{nakov2009improved} for multilingual SMT), (b) using a common subword-vocabulary across all languages \textit{e.g.} character \cite{lee17} and BPE \cite{nguyen17}, (c) representing words by  both character encoding and a latent embedding space shared
by all languages \cite{wang-ICLR2019}. 

\citet{rikters18} and \citet{lakew18} have compared RNN, CNN and the self-attention based architectures for MNMT. They show that self-attention based architectures outperform the other architectures in many cases. 


\compactpara{Minimal Sharing}. 
On the other hand, \citet{firat16} proposed a model comprised of separate embeddings, encoders and decoders for each language. By sharing attention across languages, they show improvements over bilingual models. However, this model has a large number of parameters. Nevertheless, the number of parameters only grows linearly with the number of languages, while it grows quadratically for bilingual systems spanning all the language pairs in the multiway system. 

\vspace{-2mm}
\subsection{Controlling Parameter Sharing}
\vspace{-1mm}
In between the extremities of parameter sharing exemplified by the above mentioned models, lies an array of choices. The degree of parameter sharing depends on the divergence between the languages involved \cite{sachan18} and can be controlled at various layers of the MNMT system. Sharing encoders among multiple languages is very effective and is widely used \cite{lee17,sachan18}. \citet{blackwood18} explored target language, source language and pair specific attention parameters. They showed that target language specific attention performs better than other attention sharing configurations. For self-attention based NMT models, \citet{sachan18} explored various parameter sharing strategies. They showed that sharing the decoder self-attention and encoder-decoder inter-attention  parameters is useful for linguistically dissimilar languages. \citet{Zaremoodi-ACL2018} further proposed a routing network to dynamically control parameter sharing learned from the data. Designing the right sharing strategy is important to maintaining a balance between model compactness and translation accuracy. 

\compactpara{Dynamic Parameter or Representation Generation.} Instead of defining the parameter sharing protocol a priori, \citet{platanios18} learned the degree of parameter sharing from the data. This is achieved by defining the language specific model parameters as a function of global parameters and language embeddings. This approach also reduces the number of language specific parameters (only language embeddings), while still allowing each language to have its own unique parameters for different network layers. In fact, the number of parameters is only a small multiple of the compact model (the multiplication factor accounts for the language embedding size) \cite{johnson17}, but the language embeddings can directly impact the model parameters instead of the weak influence that language tags have.

\compactpara{Universal Encoder Representation.} Ideally, multiway systems should generate encoder representations that are language agnostic. However, the attention mechanism sees a variable number of encoder representations depending on the sentence length (this could vary for translations of the same sentence). To overcome this, an attention bridge network generates a fixed number of contextual representations that are input to the attention network \cite{lu18,vazquez18}. \citet{rudramurthy19} pointed out that the contextualized embeddings are word order dependent, hence not language agnostic.

\compactpara{Multiple Target Languages.} This is a challenging scenario because parameter sharing has to be balanced with the capability to generate sentences in each target language. \citet{blackwood18} added the language tag to the beginning as well as end of sequence to avoid its attenuation in a left-to-right encoder. \citet{wang18} explored multiple methods for supporting target languages: (a) target language tag at beginning of the decoder, (b) target language dependent positional embeddings, and (c) divide  hidden units of each decoder layer into shared and language-dependent ones. Each of these methods provide gains over \citet{johnson17}, and combining all gave the best results.

\vspace{-2mm}
\subsection{Training Protocols}
\vspace{-1mm}

\compactpara{Joint Training.} All the available languages pairs are trained jointly to minimize the mean negative log-likelihood for each language pair. As some language pairs would have more data than other languages, the model may be biased. To avoid this, sentence pairs from different language pairs are sampled to maintain a healthy balance. Mini-batches can be comprised of a mix of samples from different language pairs \cite{johnson17} or the training schedule can cycle through mini-batches consisting of a language pair only \cite{firat16}. 
For architectures with language specific layers, the latter approach is convenient to implement.

\compactpara{Knowledge Distillation.} In this approach suggested by \citet{tan18}, bilingual models are first trained for all language pairs involved. These bilingual models are used as \textit{teacher models} to train a single \textit{student model} for all language pairs. The student model is trained using a linear interpolation of the standard likelihood loss as well as distillation loss that captures the distance between the output distributions of the student and teacher models. The distillation loss is applied for a language pair only if the teacher model shows better translation accuracy than the student model on the validation set. This approach shows better results than joint training of a black-box model, but training time increases significantly because bilingual models also have to be trained.

\section{Low or Zero-Resource MNMT}
\label{sec:low}
An important motivation for MNMT is to improve or support translation for language pairs with scarce or no parallel corpora, by utilizing training data from high-resource language pairs. In this section, we will discuss the MNMT approaches that specifically address the low or zero-resource scenario. 



\vspace{-2mm}
\subsection{Transfer Learning}
\label{sec:trasfer}
\vspace{-1mm}
Transfer learning \cite{Pan:2010:STL:1850483.1850545} has been widely explored to address low-resource translation, where knowledge learned from a high-resource language pair is used to improve the NMT performance on a low-resource pair.

\compactpara{Training.} Most studies have explored the following setting: the high-resource and low-resource language pairs share the same target language. \citet{DBLP:conf/emnlp/ZophYMK16:original} first showed that transfer learning can benefit low-resource language pairs. First, they trained a \textit{parent model} on a high-resource language pair. The \textit{child model} is initialized with the parent's parameters wherever possible and trained on the small parallel corpus for the low-resource pair. This process is known as \textit{fine-tuning}. They also studied the effect of fine-tuning only a subset of the child model's parameters (source and target embeddings, RNN layers and attention). The initialization has a strong regularization effect in training the child model. \citet{gu18b} used the model agnostic meta learning (MAML) framework \cite{finn17} to learn  appropriate parameter initialization from the parent pair(s) by taking the child pair into consideration. Instead of fine-tuning, both language pairs can also be jointly trained \cite{gu18}. 

\compactpara{Language Relatedness.}  \citet{DBLP:conf/emnlp/ZophYMK16:original} and \citet{Y17-1038} have empirically shown that {\it language relatedness} between the parent and child source languages has a big impact on the possible gains from transfer learning. \citet{kocmi-bojar:2018:WMT} showed that transfer learning improves low-resource language translation, even when neither the source nor the target languages are shared between the resource-rich and poor language pairs. Further investigation is needed to understand the gains in translation quality in this scenario. \citet{D18-1103} used language relatedness to prevent overfitting when rapidly adapting pre-trained MNMT model for low-resource scenarios. \citet{DBLP:journals/corr/Chaudhary19} used this approach to translate 1,095 languages to English. 

\compactpara{Lexical Transfer.} \citet{DBLP:conf/emnlp/ZophYMK16:original} randomly initialized the word embeddings of the child source language, because those could not be transferred from the parent. \citet{gu18} improved on this simple initialization by mapping pre-trained monolingual embeddings of the parent and child sources to a common vector space. On the other hand, \citet{nguyen17} utilized the lexical similarity between related source languages using a small subword vocabulary. \citet{DBLP:journals/corr/abs-1811-01137} dynamically updated the {\it vocabulary} of the parent model with the low-resource language pair before transferring parameters.

\compactpara{Syntactic Transfer.} \citet{gu18} proposed to encourage better transfer of contextual representations from parents using a mixture of language experts network. \citet{rudramurthy19} showed that reducing the {\it word order divergence} between source languages via pre-ordering is beneficial in extremely low-resource scenarios.  



\vspace{-2mm}
\subsection{Pivoting} \label{sec:pivoting}
\vspace{-1mm}
Zero-resource NMT was first explored by \citet{firat16}, where a multiway NMT model was used to translate from Spanish to French using English as a pivot language. This pivoting was done either at run time or during pre-training. 

\compactpara{Run-Time Pivoting.} \citet{firat16} involved a {\it pipeline} through paths in the multiway model, which first translates from French to English and then from English to Spanish. 
They also experimented with using the intermediate English translation as an additional source for the second stage.

\compactpara{Pivoting during Pre-Training.} \citet{firat16b} used the MNMT model to first translate the Spanish side of the training corpus to English which in turn is translated into French. This gives a pseudo-parallel French-Spanish corpus where the source is synthetic and the target is original. The MNMT model is {\it fine tuned on this synthetic data} and this enables direct French to Spanish translation. \citet{firat16b} also showed that a small clean parallel corpus between French and Spanish can be used for fine tuning and can have the same effect as a pseduo-parallel corpus which is two orders of magnitude larger.
Pivoting models can be improved if they are jointly trained as shown by \citet{ijcai2017-555}. {\it Joint training} was achieved by either forcing the pivot language's embeddings to be similar or maximizing the likelihood
of the cascaded model on a small source-target parallel
corpus.
\citet{P17-1176} proposed {\it teacher-student learning} for pivoting where they first trained a pivot-target NMT model and used it as a teacher to guide the behaviour of a source-target NMT model.

\vspace{-2mm}
\subsection{Zero-Shot} \label{sec:zeroshot}
\vspace{-1mm}
The approaches proposed so far involve pivoting or synthetic corpus generation, which is a slow process due to its two-step nature. It is more interesting, and challenging, to enable translation between a zero-resource pair without explicitly involving a pivot language during decoding or for generating pseudo-parallel corpora. 
This scenario is known as {\it zero-shot} NMT.
Zero-shot NMT also requires a pivot language but it is only used during training without the need to generate pseudo-parallel corpora.

\compactpara{Training.} Zero-shot NMT was first demonstrated by \citet{johnson17}. However, this zero-shot translation method is inferior to pivoting. They showed that the context vectors (from attention) for  unseen language pairs differ from the seen language pairs, possibly explaining the degradation in translation quality.
\citet{DBLP:journals/corr/abs-1811-01389} tried to overcome this limitation by augmenting the training data with the pseudo-parallel unseen pairs generated by iterative application of the same zero-shot translation. 
\citet{arivazhagan2018missing} included  explicit language invariance losses in the optimization function to encourage parallel sentences to have the same representation. 
{\it Reinforcement learning} for zero-shot learning was explored by \citet{DBLP:journals/corr/abs-1805-10338} where the dual learning framework was combined with rewards from language models.

\compactpara{Corpus Size.}
Work on translation for Indian languages showed that zero-shot works well only when the training corpora are extremely large \cite{Giulia-MTS2017}. As the corpora for most Indian languages contain fewer than 100k sentences, the zero-shot approach is rather infeasible despite linguistic similarity. \citet{DBLP:journals/corr/abs-1811-01389} confirmed this in the case of European languages where small training corpora were used. \citet{Giulia-MTS2017} also showed that zero-shot translation works well only when the training corpora are large, while \citet{aharoni19} show that massively multilingual models are beneficial for zeroshot translation. 

\compactpara{Language Control.} Zero-shot NMT tends to translate into the wrong language at times and \citet{DBLP:journals/corr/abs-1711-07893} proposed to filter the output of the softmax so as to force the model to translate into the desired language.

\section{Multi-Source NMT}
\label{sec:multisource}

If the same source sentence is available in multiple languages then these sentences can be used together to improve the translation into the target language. This technique is known as multi-source MT \cite{och2001statistical}.
Approaches for multi-source NMT can be extremely useful for creating N-lingual (N $>$ 3) corpora such as Europarl \cite{koehn2005epc} and UN \cite{ZIEMSKI16.1195}.
The underlying principle is to leverage redundancy in terms of source side linguistic phenomena expressed in multiple languages. 

\compactpara{Multi-Source Available.}
Most studies assume that the same sentence is available in multiple languages.
\citet{N16-1004} showed that a multi-source NMT model using {\it separate encoders and attention networks} for each source language outperforms single source models.
A simpler approach concatenated multiple {\it source sentences} and fed them to a standard NMT model \citet{Dabre-MTS2017}, with performance comparable to \cite{N16-1004}.  
Interestingly, this model could automatically identify the boundaries between different source languages and simplify the training process for multi-source NMT. \citet{Dabre-MTS2017} also showed that it is better to use linguistically similar source languages, especially in low-resource scenarios. {\it Ensembling} of individual source-target models is another beneficial  approach, for which \citet{C16-1133} proposed several methods with different degrees of parameterization.

\compactpara{Missing Source Sentences.}
There can be missing source sentences in multi-source corpora. \citet{W18-2711} extended \cite{N16-1004} by representing each ``missing'' source language with a {\it dummy token}. 
\citet{choi2018improving} and \citet{nishimura18iwslt} further proposed to use MT generated {\it synthetic sentences}, instead of a dummy token for the missing source languages. 

\compactpara{Post-Editing.} Instead of having a translator translate from scratch, multi-source NMT can be used to generate high quality translations. The translations can then be post-edited, a process that is less labor intensive and cheaper compared to translating from scratch. Multi-source NMT has been used for post-editing where the translated sentence is used as an additional source, leading to improvements \cite{W17-4773}.



\section{Multilingualism in Older Paradigms}
\label{sec:history}
One of the long term goals of the MT community is the development of architectures that can handle more than two languages. 

\paragraph{RBMT.} To this end, rule-based systems (RBMT) using an \textit{interlingua} were explored widely in the past. The interlingua is a symbolic semantic, language-independent representation for natural language text \cite{Sgall:1987:MTL:976858.976876}. Two popular interlinguas are UNL \cite{uchida1996unl} and AMR \cite{banarescu-EtAl:2013:LAW7-ID} Different interlinguas have been proposed in various systems like KANT \cite{nyberg1997kant}, UNL, UNITRAN \cite{dorr1987unitran} and DLT \cite{witkam2006dlt}. Language specific analyzers converted language input to interlingua, while language specific decoders converted the interlingua into another language. To achieve an unambiguous semantic representation, a lot of linguistic analysis had to be performed and many linguistic resources were required. Hence, in practice, most interlingua systems were limited to research systems or translation in specific domains and could not scale to many languages. Over time most MT research focused on building bilingual systems. 

\paragraph{SMT.} Phrase-based SMT (PBSMT) systems \cite{koehn2003statistical}, a very successful MT paradigm, were also bilingual for the most part. Compared to RBMT, PBSMT requires less linguistic resources and instead requires parallel corpora. However, like RBMT, they work with symbolic, discrete representations making multilingual representation difficult. Moreover, the central unit in PBSMT is the \textit{phrase}, an ordered sequence of words (not in the linguistic sense). Given its arbitrary structure, it is not clear how to build a common symbolic representation for phrases across languages. Nevertheless, some shallow forms of multilingualism have been explored in the context of: (a) pivot-based SMT, (b) multi-source PBSMT, and (c) SMT involving related languages. 

\noindent\textit{Pivoting.} Popular solutions are: chaining source-pivot and pivot-target systems at decoding \cite{utiyama2007}, training a source-target system using synthetic data generated using target-pivot and pivot-source systems \cite{de2006catalanenglish}, and phrase-table triangulation pivoting source-pivot and pivot-target phrase tables \cite{utiyama2007,wu2007pivot}. 

\noindent\textit{Multi-source.}  Typical approaches are: re-ranking outputs from independent source-target systems \cite{och2001statistical}, composing a new output from independent source-target outputs \cite{matusov2006computing}, and translating a combined input representation of multiple sources using lattice networks over multiple phrase tables \cite{schroeder2009word}. 


\noindent\textit{Related languages.} For multilingual translation with multiple related source languages, the typical approaches involved script unification by mapping to a common script such as Devanagari \cite{banerjee2018multilingual} or transliteration \cite{nakov2009improved}. Lexical similarity was utilized using subword-level translation models \cite{vilar2007can,tiedemann2012character,kunchukuttan2016orthographic,kunchukuttan2017bpe}. Combining subword-level representation and pivoting for translation among related languages has been explored \citep{henriquez2011pivot,tiedemann2012character,kunchukuttan2017pivot}. Most of the above mentioned multilingual systems involved either decoding-time operations, chaining black-box systems or composing new phrase-tables from existing ones.

\paragraph{Comparison with MNMT.}
While symbolic representations constrain a unified multilingual representation, distributed universal language representation using real-valued vector spaces makes multilingualism easier to implement in NMT. As no language specific feature engineering is required for NMT, making it possible to scale to multiple languages. Neural networks provide flexibility in experimenting with a wide variety of architectures, while advances in optimization techniques and availability of deep learning toolkits make prototyping faster. 

\section{Datasets and Resources}
\label{sec:data}

MNMT requires parallel corpora in similar domains across multiple languages. 

\compactpara{Multiway.} Commonly used publicly available multilingual parallel corpora are the TED corpus \cite{mauro2012wit3}, UN Corpus \cite{ziemski2016united} and those from the European Union like Europarl, JRC-Aquis, DGT-Aquis, DGT-TM, ECDC-TM, EAC-TM \cite{steinberger2014overview}. While these sources are primarily comprised of European languages, parallel corpora for some Asian languages is accessible through the WAT shared task \cite{nakazawa2018overview}. Only small amount of parallel corpora are available for many languages, primarily from movie subtitles and software localization strings \cite{TIEDEMANN12.463}. 

\compactpara{Low or Zero-Resource.} For low or zero-resource NMT translation tasks, good test sets are required for evaluating translation quality. The above mentioned multilingual parallel corpora can be a source for such test sets. In addition, there are other small parallel datasets like the FLORES dataset for English-\{Nepali,Sinhala\} \cite{guzman2019flores}, the XNLI test set spanning 15 languages \cite{conneau2018xnli} and the Indic parallel corpus \cite{birch2011indic}. The WMT shared tasks \cite{bojar2018wmtfindings} also provide test sets for some low-resource language pairs. 

\compactpara{Multi-Source.} The corpora for multi-source NMT have to be aligned across languages. Multi-source corpora can be extracted from some of the above mentioned sources. The following are widely used for evaluation in the literature: Europarl \cite{koehn2005epc}, TED \cite{TIEDEMANN12.463}, UN \cite{ZIEMSKI16.1195}. The Indian Language Corpora Initiative (ILCI) corpus \cite{jha2010tdil} is a 11-way parallel corpus of Indian languages along with English. The Asian Language Treebank \cite{thu2016introducing} is a 9-way
parallel corpus of South-East Asian languages along with English, Japanese and Bengali. The MMCR4NLP project \cite{dabre2017mmcr4nlp} compiles language family grouped multi-source corpora and provides standard splits.

\compactpara{Shared Tasks.} Recently, shared tasks with a focus on multilingual translation have been conducted at IWSLT \cite{cettolo2017overview}, WAT \cite{nakazawa2018overview} and WMT \cite{bojar2018wmtfindings}; so common benchmarks are available.




\section{Connections with Domain Adaptation}
\label{sec:domain}
High quality parallel corpora are limited to specific domains.
Both, vanilla SMT and NMT perform poorly for domain specific translation in low-resource scenarios \cite{duh-EtAl:2013:Short,koehn-knowles:2017:NMT}.
Leveraging out-of-domain parallel corpora and in-domain monolingual corpora for in-domain translation is known as domain adaptation for MT \cite{C18-1111}. 

As we can treat each domain as a language, there are many similarities and common approaches between MNMT and domain adaptation for NMT. 
Therefore, similar to MNMT, when using out-of-domain parallel corpora for domain adaptation, multi-domain NMT and transfer learning based approaches \cite{P17-2061} have been proposed for domain adaptation.
When using in-domain monolingual corpora, a typical way of doing domain adaptation is generating a pseduo-parallel corpus by back-translating target in-domain monolingual corpora \cite{sennrich-haddow-birch:2016:P16-11}, which is similar to the pseduo-parallel corpus generation in MNMT \cite{firat16b}.

There are also many differences between MNMT and domain adaptation for NMT. While pivoting is a popular approach for MNMT \cite{ijcai2017-555}, it is unsuitable for domain adaptation.
As there are always vocabulary overlaps between different domains, there are no zero-shot translation \cite{johnson17} settings in domain adaptation. In addition, it not uncommon to write domain specific sentences in different styles and so
multi-source approaches \cite{N16-1004} are not applicable either. 
On the other hand, data selection approaches in domain adaptation that select out-of-domain sentences which are similar to in-domain sentences \shortcite{wang-EtAl:2017:Short3} have not been applied to MNMT. In addition, instance weighting approaches \cite{iwnmt}  that interpolate in-domain and out-of-domain models have not been studied for MNMT. However, with the development of cross-lingual sentence embeddings, data selection and instance weighting approaches might be applicable for MNMT in the near future.




\vspace{-1mm}
\section{Future Research Directions}
\label{sec:future}
\vspace{-1mm}
While exciting advances have been made in MNMT in recent years, there are still many interesting directions for exploration. 

\compactpara{Language Agnostic Representation Learning.}
A core question that needs further investigation is: how do we build encoder and decoder representations that are language agnostic? Particularly, the questions of word-order divergence between the source languages and variable length encoder representations have received little attention. 

\compactpara{Multiple Target Language MNMT.} Most current efforts address multiple source languages.
Multiway systems for multiple low-resource target languages need more attention.  The right balance between sharing representations \textit{vs.} maintaining the distinctiveness of the target language for generation needs exploring. 

\compactpara{Explore Pre-training Models.}
Pre-training embeddings, encoders and decoders have been shown to be useful for NMT \cite{ramachandran17}. How pre-training can be incorporated into different MNMT architectures, is an important as well. Recent advances in cross-lingual word \cite{klementiev12a,mikolov13a,chandar14a,artetxe16a,conneau18a,jawanpuria2018learning} and sentence embeddings \cite{conneau2018xnli,chen2018multilingual,artetxe2019multilingual} could provide directions for this line of investigation.   

\compactpara{Related Languages, Language Registers and Dialects.}
Translation involving related languages, language registers and dialects can be further explored given the importance of this use case. 

\compactpara{Code-Mixed Language.}
Addressing intra-sentence multilingualism \textit{i.e.} code mixed input and output, creoles and pidgins is an interesting research direction. The compact MNMT models can handle code-mixed input, but code-mixed output remains an open problem \cite{johnson17}. 

\compactpara{Multilingual and Multi-Domain NMT.}
Jointly tackling multilingual and multi-domain translation is an interesting direction with many practical use cases. When extending an NMT system to a new language, the parallel corpus in the domain of interest may not be available. Transfer learning in this case has to span languages and domains.



\vspace{-1mm}
\section{Conclusion}
\label{sec:conlusion}
\vspace{-1mm}
MNMT has made rapid progress in the recent past. In this survey, we have covered literature pertaining to the major scenarios we identified for multilingual NMT: multiway, low or zero-resource (transfer learning, pivoting, and zero-shot approaches) and multi-source translation. We have systematically compiled the principal design approaches and their variants, central MNMT issues and their proposed solutions along with their strengths and weaknesses. We have put MNMT in a historical perspective w.r.t work on multilingual RBMT and SMT systems. We  suggest promising and important directions for future work. We hope that this survey paper could significantly promote and accelerate MNMT research.


\bibliography{acl2019_mnmt}
\bibliographystyle{acl_natbib}

\end{document}


\maketitle

\begin{table}[h]
    \centering
    \scalebox{.8}{
    \begin{tabular}{|p{15em}|p{35em}|}
    
         \hline
         \multicolumn{2}{|c|}{\textbf{Multiway}}  \\
         \hline
         \multicolumn{2}{|l|}{\textbf{Prototypical Approaches}} \\ \hline 
         Complete Sharing & \cite{johnson17},   \cite{aharoni19}, \cite{ha16}, \cite{sachan18}, \cite{dabre18wat}, \cite{lee17}, \cite{nguyen17}, \cite{wang-ICLR2019}, \cite{rikters18}, \cite{lakew18} \\       \hline
         Minimal  Sharing & \cite{firat16}\\         \hline
         \multicolumn{2}{|l|}{\textbf{Controlling Sharing}} \\ \hline
         General & \cite{sachan18}, \cite{lee17}, \cite{blackwood18}, \cite{Zaremoodi-ACL2018} \\         \hline
         Dynamic Parameter Generation & \cite{platanios18} \\         \hline
         Universal Encoder Representation & \cite{lu18}, \cite{vazquez18}, \cite{rudramurthy19} \\ \hline 
         Multiple Target Languages & \cite{blackwood18}, \cite{wang18} \\         \hline
         \multicolumn{2}{|l|}{\textbf{Training Protocols}} \\ \hline
         Joint Training & \cite{johnson17}, \cite{firat16} \\         \hline
         Knowledge Distillation & \cite{tan18} \\     
         
         \hline
         \multicolumn{2}{|c|}{\textbf{Low or Zero-Resource}}  \\
         \hline 
         \multicolumn{2}{|l|}{\textbf{Transfer Learning}}\\ \hline 
         Training & \cite{DBLP:conf/emnlp/ZophYMK16:original}, \cite{gu18}, \cite{gu18b}\\ \hline
         Language Relatedness & \cite{DBLP:conf/emnlp/ZophYMK16:original}, \cite{Y17-1038}, \cite{kocmi-bojar:2018:WMT}, \cite{D18-1103}, \cite{DBLP:journals/corr/Chaudhary19}  \\ \hline
         Lexical Transfer & \cite{DBLP:conf/emnlp/ZophYMK16:original}, \cite{gu18}, \cite{nguyen17}, \cite{DBLP:journals/corr/abs-1811-01137} \\ \hline
         Syntactic Transfer & \cite{gu18}, \cite{rudramurthy19} \\   \hline
         \multicolumn{2}{|l|}{\textbf{Pivot}} \\ \hline
         Runtime &  \cite{firat16}\\ \hline
         Pre-train & \cite{firat16b}, \cite{ijcai2017-555}, \cite{P17-1176} \\ \hline           
         \multicolumn{2}{|l|}{\textbf{Zeroshot}} \\ \hline 
         Training & \cite{johnson17}, \cite{DBLP:journals/corr/abs-1811-01389}, \cite{arivazhagan2018missing}, \cite{DBLP:journals/corr/abs-1805-10338} \\ \hline
         Corpus Size & \cite{Giulia-MTS2017}, \cite{DBLP:journals/corr/abs-1811-01389}, \cite{Giulia-MTS2017}, \cite{aharoni19} \\ \hline
         Language Control & \cite{DBLP:journals/corr/abs-1711-07893} \\ \hline
         
         \hline
         \multicolumn{2}{|c|}{\textbf{Multi-source}}  \\
         \hline
         Multi-source available & \cite{N16-1004}, \cite{Dabre-MTS2017},  \cite{C16-1133} \\         \hline
         Missing source sentences & \cite{W18-2711}, \cite{choi2018improving}, \cite{nishimura18iwslt} \\         \hline
         Post-editing & \cite{W17-4773} \\
         \hline
    \end{tabular}
    }
    \label{tab:Related Works}
    \caption{Papers related to MNMT categorized according to resource scenarios and underlying modeling principles}
    
\end{table}

\clearpage

\bibliography{acl2019_mlnmt}
\bibliographystyle{acl_natbib}


\maketitle

\begin{table}[]
    \centering
    \begin{tabular}{|l|l|}
         \hline
         \multicolumn{2}{|c|}{\textbf{Multiway}}  \\
         \hline
         \multicolumn{2}{|l|}{\textbf{Prototypical Approaches}} \\ \hline 
         Complete Sharing & \cite{ha16}, \cite{johnson17},   \cite{aharoni19}, \cite{sachan18} \\       \hline
         Minimal  Sharing & \\         \hline
         \multicolumn{2}{|l|}{\textbf{Controlling Sharing}} \\ \hline
         General & \\         \hline
         Dynamic Parameter Generation & \\         \hline
         Universal Encoder Representation & \\ \hline 
         Multiple Target Languages & \\         \hline
         \multicolumn{2}{|l|}{\textbf{Training Protocols}} \\ \hline
         Joint Training & \\         \hline
         Knowledge Distillation & \\        
         \hline
         \multicolumn{2}{|c|}{\textbf{Low or Zero-Resource}}  \\
         \hline 
         \multicolumn{2}{|l|}{\textbf{Transfer Learning}}\\ \hline 
         \multicolumn{2}{|l|}{\textbf{Pivot}} \\ \hline
         \multicolumn{2}{|l|}{\textbf{Zeroshot}} \\ \hline 
         
         \hline
         \multicolumn{2}{|c|}{\textbf{Multi-source}}  \\
         \hline
         Multi-source available & \\         \hline
         Missing source sentences & \\         \hline
         Post-editing & \\
         \hline
    \end{tabular}
    \caption{Caption}
    \label{tab:Related Works}
\end{table}

\bibliography{acl2019_mnmt}
\bibliographystyle{acl_natbib}